\pdfoutput=1

\documentclass[11pt]{article}

\usepackage{emnlp2021}

\usepackage{times}
\usepackage{latexsym}

\usepackage[T1]{fontenc}
\usepackage{todonotes}

\usepackage[utf8]{inputenc}

\usepackage{microtype}

\title{Explaining Classes through Word Attributions}

\author{Samuel R{\"o}nnqvist, Amanda Myntti, Aki-Juhani Kyr{\"o}l{\"a}inen \\
\textbf{Sampo Pyysalo, Veronika Laippala} and \textbf{Filip Ginter} \\
TurkuNLP \\ University of Turku, Finland \\ \texttt{first.last@utu.fi}}

\begin{document}
\maketitle

\section{Introduction}

We propose a method for explaining classes in text classification tasks using deep learning models and feature attribution techniques, such as the Integrated Gradients (IG) method introduced by \citet{sundararajan2017axiomatic}. We focus specifically on IG as it provides a general framework for estimating feature importance in deep neural networks and has been shown to provide reliable saliency maps in text classification tasks among others \cite{bastings2020elephant,kokhlikyan2020captum}. 

Recently, explaining the predictions of deep neural networks has attracted a considerable amount of research interest in fields such as NLP and computer vision. Given the importance of this endeavour, several different techniques have been suggested in order to interpret model predictions \cite[see][for recent discussion]{montavon2018}. Nevertheless, these tend to focus on explaining individual predictions rather than how models perceive whole classes. To the best of our knowledge, we present the first method for aggregating explanations of individual examples in text classification to general descriptions of the classes. The method consists of three steps: 1) repeated model training and application of IG on random train/test splits, 2) aggregation of word scores of individual examples and extraction of keywords, and 3) filtering to remove spurious keywords.

We test this method by training Transformer-based text classifiers on a large Web register identification corpus and show that it is able to provide descriptive keywords for the classes. The class descriptions provide both linguistic insight and a means for analyzing and debugging neural classification models in text classification.



\section{Data and classifier} 

In our experiments, we focus on text classification using the Corpus of Online Registers of English (CORE) \cite{egbert:core}, a large-scale collection of Web texts annotated for their \textit{register} (genre) \cite{biber88}. The CORE registers are coded using a two-level taxonomy. In this study, we focus on the upper level which consists of eight register classes: Narrative (NA), Opinion (OP), How-to (HI), Interactive discussion (ID), Informational description (IN), Lyrical (LY), Spoken (SP) and Informational persuasion (IP). The dataset features the full range of registers found on the unrestricted open Web and consists of nearly 50,000 texts. In our experiments, we combine the train and development sets, totaling 38,760 documents. 

Web registers have been frequently studied in recent research both in linguistics and NLP \cite{tiktak2013,dayter-reddit-2021, vidulin2019,biber-egbert-registervariationonline2019}. 
The range of linguistic variation has, however, caused challenges for both fields, and, in particular, Web register identification studies have lacked robustness \cite{sharoff-etal-2010-web, Petrenz2011}. The method we propose in this study can benefit both fields as it provides insight about classification models and the corpora they are trained on, including potential biases. 

As a classifier, we use the XLM-R deep language model \cite{conneau-etal-2020-unsupervised} because of its strong ability to model multiple languages, both in monolingual and cross-lingual settings. We use the base size, since it uses less resources and its predictive performance on the CORE corpus is competitive with XLM-R large \cite{repo2021zeroshot}.
The task is modeled as a multilabel classification task.



\section{Method}
The descriptions of classes are extracted through the following steps:

\textbf{Step 1: Train and explain.} We combine the training and development sets of the corpus and randomly split them into a new training and validation set according to a set ratio $r$, using stratification to keep class distributions stable \cite[cf.][]{Laippala2021}.
The pre-trained language model is loaded and the decision layer (a sequence regression head) is randomly initialized. Both are fine-tuned on the new training set. Text examples in the validation set are classified and the IG method is applied in order to obtain attribution scores for the network inputs, i.e., each dimension of each input token embedding, w.r.t. each predicted class $c$. The embedding dimensions are summed up per token to provide a token-level score and all tokens in a document $d$ are normalized by the L$_2$ norm. This provides a word attribution score $s_{w,d,c}$ directly if the word $w$ consists of a single token, otherwise it is calculated as the maximum of all sub-word token scores. 

\textbf{Step 2: Aggregate attributions.} We calculate the average attribution scores $\bar{s}_{w,c}$, for each $(w,c)$, as a means for ranking of keywords per class. In order to reduce noise, we only select the $n$ top-scoring words per document $d$, and we only consider true positive predictions. We note that the method could alternatively be used for error analysis by targeting false predictions.

\textbf{Step 3: Select stable keywords.} The above process is repeated $N$ times, each time randomly shuffling and splitting the data according to Step 1, in order to quantify the stability of the keywords.
The keyword candidates ranked by $\bar{s}_{w,c}$ are filtered based on selection frequency: a word is considered stable if the ratio by which it is selected (in Step 2) across the experiments is larger than a threshold value $t$.
We also ignore words that occur in $k$ documents or less in the corpus.

The selection frequency filtering allows us to remove keywords that are unstable across runs, likely reflecting spurious features, for instance, resulting from an unrepresentative split of the data or stochastic factors in the training of the classifier itself. \citet{mccoy2020berts} show in repeated experiments on a text inference task with random initialization of the decision layer and randomized order of training examples that, while consistent test set performance was achieved, the degree of generalization as measured on a related task varied significantly. Similarly, we test the persistence and presumed generalizability of the estimated keywords by considering the randomness both in training and in data selection.

In our experiments, we have used the parameters $r=0.67$, $n=20$, $N=100$, $t=0.6$ and $k=5$.



\section{Results}

The classifiers trained in our 100 experiments achieved a mean micro average F1-score of 65.10\% ($SD=6.72\%$) and mean class-wise F1-scores in the range 26.45\%--82.92\% for the eight main register classes (see Table~\ref{tab:pred_scores} in Appendix). The Spoken (SP) class stands out as a particularly difficult case where performance was particularly unstable ($SD=27.09\%$), partly due to its small size.

Our method was able to produce descriptive keywords that clearly reflect our understanding of all the main classes (see Table~\ref{tab:keywords} in Appendix) except for the Spoken class, where no keyword surpassed the selection frequency threshold. The keywords reflect both topical and functional features typical of the registers. For instance, the highest scoring words for Interactive discussion (ID) were \textit{question, faq, forum, answer}.
Similarly, we observe other register-specific linguistic characteristics, such as words associated with research papers in Informational description (IN) and with news in Narrative (NA). The keywords also share many similarities with keywords produced with other methods applied in previous studies \cite[e.g.,][]{biber-egbert-registervariationonline2019,Laippala2021}.

Furthermore, the estimated keywords display a strong discriminative power as indicated by their uniqueness in the respective register classes.
On average, 82 ($SD=4.6$) of the top 100 keywords for a given register were not shared with the other registers demonstrating that the method was able to identify register-specific keywords. Moreover, the selection frequency of the keywords across the 100 rounds demonstrated their stability -- they are consistently identified, often in over 90\% of the repetitions. 



\section{Conclusion}
We have proposed a method for describing classes in a text classification task based on IG attributions on predictions and shown that it produces stable and interpretable results for Web register classification with XLM-R. We see the method as generally applicable and useful for studying text classes also beyond registers.
In the future, we seek to extend the method and its evaluation, and apply the approach to other languages and cross-lingual settings. In particular, the comparison of monolingual and zero-shot models will be informative of both the linguistic characteristics of registers and what models such as XLM-R learn to recognize.


\bibliography{anthology,custom}
\bibliographystyle{acl_natbib}

\appendix
\section*{Appendix}
\label{sec:appendix}

\begin{table}[h]
    \centering
    \small
    \begin{tabular}{l|rrr}
\hline
Class & F1 (\emph{M}) & \emph{SD} & Sup. (\emph{M}) \\\hline
Lyrical (LY)	&0.8292	&0.0866	&172 \\
Narrative (NA)	&0.7870	&0.0795	&5775 \\
Inter. discussion (ID)	&0.7623	&0.0787	&876 \\
Inform. description (IN)	&0.6336	&0.0662	&3399 \\
How-to (HI)	&0.5515	&0.0719	&521 \\
Opinion (OP)	&0.5379	&0.0839	&2854 \\
Inform. persuasion (IP)	&0.4094	&0.0573	&531 \\
Spoken (SP)	&0.2645	&0.2709	&206 \\\hline
Micro AVG &	0.6510	&0.0672	& -- \\ \hline
    \end{tabular}
    \caption{Predictive performance of models (N=100).}
    \label{tab:pred_scores}
\end{table}

\begin{table*}[]
    \centering
    \small
    \begin{tabular}{l|ll}
\multicolumn{3}{c}{--- How-to (HI) ---} \\
Word & Score & SF(\%) \\\hline
        how & 0.4820 & 100  \\ 
        recipe & 0.3439 & 100 \\ 
        recipes & 0.3356 & 100 \\ 
        tips & 0.3224 & 100  \\ 
        scenario & 0.3184 & 67  \\ 
        tricks & 0.2883 & 71  \\ 
        tutorial & 0.2485 & 100  \\ 
        taking & 0.2458 & 70 \\ 
        flavor & 0.2427 & 78 \\ 
        ingredients & 0.2355 & 100 \\ 
        ways & 0.2337 & 98  \\ 
        diy & 0.2307 & 83  \\ 
        associated & 0.2299 & 77 \\ 
        to & 0.2276 & 100  \\ 
        picking & 0.2254 & 86 \\ 
\multicolumn{3}{c}{}
    \end{tabular}
    \begin{tabular}{l|ll}
\multicolumn{3}{c}{--- Inter. Discussion (ID) ---} \\
    Word & Score & SF(\%) \\\hline
        question & 0.5874 & 100 \\ 
        faq & 0.5818 & 99 \\ 
        forum & 0.4855 & 100 \\ 
        answer & 0.4799 & 100 \\ 
        answers & 0.4636 & 100 \\ 
        answered & 0.4524 & 100 \\ 
        forums & 0.4232 & 100 \\ 
        replies & 0.4028 & 99 \\ 
        thread & 0.3975 & 100 \\ 
        re & 0.3833 & 100 \\ 
        discuss & 0.3363 & 100 \\ 
        threads & 0.3155 & 100 \\ 
        hello & 0.3102 & 98 \\ 
        quote & 0.3067 & 100 \\ 
        imo & 0.2988 & 99 \\ 
\multicolumn{3}{c}{}
    \end{tabular}
    \begin{tabular}{l|ll}
\multicolumn{3}{c}{--- Inform. Description (IN) ---} \\
    Word & Score & SF(\%) \\\hline
            abstract & 0.6054 & 100 \\ 
        geoscience & 0.4558 & 97 \\ 
        faqs & 0.4051 & 100  \\ 
        faq & 0.3929 & 96 \\ 
        analysing & 0.3679 & 77 \\ 
        storyline & 0.3662 & 99 \\ 
        downloads & 0.3628 & 98 \\ 
        abstracts & 0.3594 & 98 \\ 
        hal & 0.3495 & 69 \\ 
        aspect & 0.3388 & 99 \\ 
        wikis & 0.3289 & 70 \\ 
        economical & 0.3162 & 90 \\ 
        demographics & 0.3118 & 100 \\ 
        introduction & 0.2931 & 100  \\ 
        moscow & 0.2897 & 65 \\ 
\multicolumn{3}{c}{}
    \end{tabular}
    \begin{tabular}{l|ll}
\multicolumn{3}{c}{--- Inform. Persuasion (IP) ---} \\
    Word & Score & SF(\%) \\\hline
            description & 0.4922 & 100 \\ 
        pdf & 0.4031 & 73 \\ 
        publishers & 0.3934 & 67 \\ 
        isbn & 0.3821 & 98 \\ 
        discounts & 0.3644 & 76 \\ 
        rates & 0.3065 & 82 \\ 
        deal & 0.2953 & 74 \\ 
        book & 0.2805 & 100 \\ 
        relax & 0.2635 & 74 \\ 
        editions & 0.2555 & 75 \\ 
        luxury & 0.2512 & 93 \\ 
        rental & 0.2472 & 97 \\ 
        shop & 0.2464 & 99 \\ 
        stylish & 0.2418 & 87 \\ 
        prices & 0.2358 & 83 \\ 
\multicolumn{3}{c}{}
    \end{tabular}
    \begin{tabular}{l|ll}
\multicolumn{3}{c}{--- Lyrical (LY) ---} \\
    Word & Score & SF(\%) \\\hline
            lyrics & 0.3772 & 100 \\ 
        music & 0.2891 & 93 \\ 
        poem & 0.2511 & 94 \\ 
        comment & 0.2148 & 70 \\ 
        chords & 0.1893 & 82 \\ 
        hate & 0.1795 & 75 \\ 
        guitar & 0.1794 & 81 \\ 
        truth & 0.1710 & 98 \\ 
        finally & 0.1640 & 90 \\ 
        thanks & 0.1622 & 79 \\ 
        chorus & 0.1597 & 66 \\ 
        happiness & 0.1570 & 83 \\ 
        stood & 0.1554 & 89 \\ 
        album & 0.1551 & 63 \\ 
        gotta & 0.1494 & 98 \\ 
\multicolumn{3}{c}{}
    \end{tabular}
\begin{tabular}{l|ll}
\multicolumn{3}{c}{--- Narrative (NA) ---} \\
    Word & Score & SF(\%) \\\hline
            newswire & 0.5669 & 100 \\ 
        reddit & 0.4565 & 100 \\ 
        afp & 0.4212 & 100 \\ 
        ufc & 0.3976 & 100 \\ 
        bundesliga & 0.3803 & 100 \\ 
        flickr & 0.3736 & 100 \\ 
        kardashians & 0.3720 & 76 \\ 
        reuters & 0.3618 & 100 \\ 
        1867 & 0.3614 & 92  \\ 
        nba & 0.3587 & 100 \\ 
        lollies & 0.3519 & 66 \\ 
        blogosphere & 0.3511 & 100 \\ 
        gmt & 0.3389 & 100 \\ 
        gutted & 0.3378 & 96 \\ 
        playoffs & 0.3328 & 100 \\ 
\multicolumn{3}{c}{}
    \end{tabular}
    
    \begin{tabular}{l|ll}
\multicolumn{3}{c}{--- Opinion (OP) ---} \\
    Word & Score & SF(\%) \\\hline
            psalms & 0.7098 & 94 \\ 
        weblog & 0.5511 & 91  \\ 
        review & 0.5355 & 100 \\ 
        psalm & 0.4798 & 100  \\ 
        forbes & 0.4506 & 76  \\ 
        horrors & 0.3883 & 82  \\ 
        blog & 0.3705 & 100  \\ 
        blogged & 0.3625 & 85 \\ 
        disclaimer & 0.3597 & 85 \\ 
        categories & 0.3568 & 72 \\ 
        evaluating & 0.3560 & 62 \\ 
        poll & 0.3517 & 83 \\ 
        monday & 0.3446 & 100 \\ 
        tips & 0.3437 & 100 \\ 
        jeremiah & 0.3418 & 96 \\ 
\multicolumn{3}{c}{}
    \end{tabular}
    \caption{Top-15 extracted keywords for each register class ranked by mean aggregated attribution score (Score). The lists are filtered by threshold on selection frequency (SF).
    }
    \label{tab:keywords}
\end{table*}

\end{document}